\documentclass[10pt,twocolumn,letterpaper]{article}

\usepackage{cvpr}
\usepackage{times}
\usepackage{epsfig}
\usepackage{graphicx}
\usepackage{amsmath}
\usepackage{amssymb}
\usepackage{caption}
\usepackage{bm}
\usepackage{multirow}
\usepackage{cuted}
\usepackage{capt-of}
\usepackage[ruled,vlined]{algorithm2e}
\usepackage{algpseudocode}  
\usepackage{pdfpages}

\usepackage[pagebackref=true,breaklinks=true,letterpaper=true,colorlinks,bookmarks=false]{hyperref}

\cvprfinalcopy 

\def\cvprPaperID{564} 

\ifcvprfinal\fi
\begin{document}
	
\graphicspath{{images_clean/}}

\newcommand{\bx}{{\bf x}}
\newcommand{\bz}{{\bf z}}
\newcommand{\balpha}{{\bm \alpha}}
\newcommand{\bbeta}{{\bm \beta}}
\newcommand{\bdelta}{{\bm \delta}}
\newcommand{\bgamma}{{\bm \gamma}}
\newcommand{\btheta}{{\bm \theta}}
\newcommand{\blambda}{{\bm \lambda}}
\newcommand{\bS}{{\bf S}}
\newcommand{\bT}{{\bf T}}
\newcommand{\bB}{{\bf B}}
\newcommand{\tmp}{u}

\title{Disentangled and Controllable Face Image Generation via 3D Imitative-Contrastive Learning}

\author{Yu Deng\thanks{This work was done when Yu Deng was an intern at MSRA.}\,\,$^{1,2}$ \quad Jiaolong Yang$^{2}$ \quad Dong Chen$^{2}$ \quad Fang Wen$^{2}$ \quad Xin Tong$^{2}$ \\
	$^1${Tsinghua University} \quad  $^2${Microsoft Research Asia} \\
	{\tt\small \{t-yudeng,jiaoyan,doch,fangwen,xtong\}@microsoft.com}
}

\maketitle
\begin{strip}
	\vspace{-56pt}
	\centering
	\includegraphics[width=\textwidth]{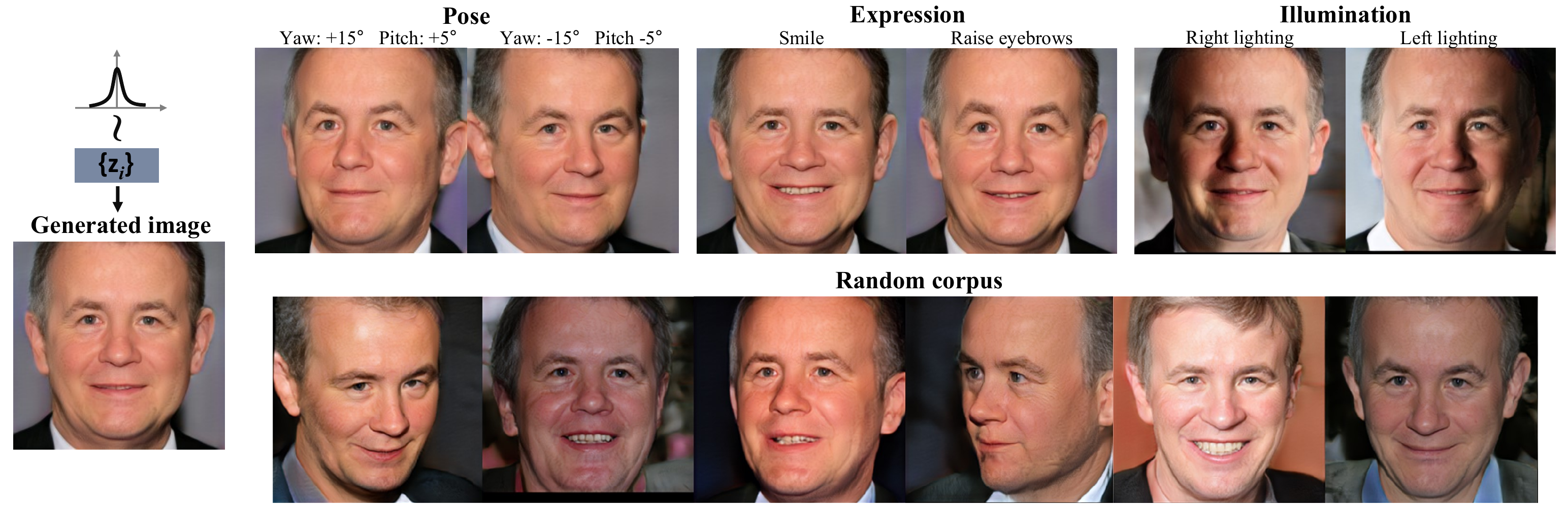}
	\vspace{-21pt}
	\captionof{figure}{This paper presents \textit{\textbf{DiscoFaceGAN}} that generates realistic face images of virtual people with independent latent variables of identity, expression, pose, and illumination.  The latent space is interpretable and highly disentangled, which allows precise control of the targeted images (\eg, degree of each pose angle, lighting intensity and direction), as shown in the top row. The bottom row shows the generated images when we keep the identity and randomize other properties. The faces generated by our method are not any real person in the world.\label{fig:teaser}}
	\vspace{-6pt}
\end{strip}

\begin{abstract}
	\vspace{-3pt}
	We propose \textbf{DiscoFaceGAN}, an approach for face image generation with \textbf{DIS}entangled, precisely-\textbf{CO}ntrollable latent representations for identity of non-existing people, expression, pose, and illumination. We embed 3D priors into adversarial learning and train the network to imitate the image formation of an analytic 3D face deformation and rendering process. To deal with the generation freedom induced by the domain gap between real and rendered faces, we further introduce contrastive learning to promote disentanglement by comparing pairs of generated images. Experiments show that through our imitative-contrastive learning, the factor variations are very well disentangled and the properties of a generated face can be precisely controlled. 
	We also analyze the learned latent space and present several meaningful properties supporting factor disentanglement. Our method can also be used to embed real images into the disentangled latent space.
	We hope our method could provide new understandings of the relationship between physical properties and deep image synthesis. \footnote{Code available \href{https://github.com/microsoft/DisentangledFaceGAN}{here}.}
\end{abstract}

\vspace{-17pt}
\section{Introduction}
Face image synthesis has achieved tremendous success in the past few years with the rapid advance of Generative Adversarial Networks (GANs)~\cite{goodfellow2014generative}.
State-of-the art GAN models, such as the recent StyleGAN~\cite{karras2019style}, can generate high-fidelity virtual face images that are sometimes even hard to distinguish from real ones.

Compared to the vast body of works devoted to improving the image generation quality and tailoring GANs for various applications, synthesizing face images \emph{de novo} with multiple disentangled latent spaces characterizing different properties of a face image is still not well investigated. 
Such a disentangled latent representation is desirable for constrained face image generation (\eg, random
identities with specific illuminations or poses). It can also derive a disentangled representation of a real image by embedding it into the learned feature space.
A seminal GAN research for disentangled image generation is InfoGAN~\cite{chen2016infogan}, 
where the representation disentanglement is learned in an unsupervised manner via maximizing the mutual information between the latent variables and the observation.
However, it has been shown that without any prior or weak supervision, there is no guarantee that each latent variable contains 
a semantically-meaningful factor of variation~\cite{pmlr-v97-locatello19a,dai2019diagnosing}.

In this paper, we investigate synthesizing face images of virtual people 
with independent latent variables for identity, expression, pose, lighting, and an additional noise. To gain predictable controllability on the former four variables, we translate them to the coefficients of parametric models
through training a set of Variational Autoencourders (VAE). We incorporate priors from 3D Morphable Face Models (3DMM)~\cite{blanz1999morphable,paysan20093d} and an analytic rendering procedure into adversarial learning. A set of \emph{imitative losses} is introduced which enforces the generator to imitate the explainable image rendering process, thus generating face properties characterized by the latent variables. However, the domain gap between real and rendered faces gives rise to a certain generation freedom that is uncontrollable, leading to unsatisfactory disentanglement of factor variations.

To deal with such generation freedom and enhance disentanglement, 
we further propose a collection of \emph{contrastive losses} for training. We compare pairs of generated images and penalize the appearance difference that is only induced by a set of identical latent variables shared between each pair. This way, the generator is forced to express an independent influence of each latent variable to the final output. We show that these contrastive losses are crucial to achieve complete latent variable disentanglement.

The model we use in this paper is based on the StyleGAN structure~\cite{karras2019style}, though our method can be extended to other GAN models as well. We modify the latent code layer of StyleGAN and equip it with our new loss functions for training. We show that the latent variables can be highly disentangled and the generation can be accurately controlled. Similar to StyleGAN, the faces generated by our method do not correspond to any real person in the world. We further analyze the learned StyleGAN latent space and find some meaningful properties supporting factor disentanglement. Our method can be used to embed real images into the disentangled latent space and we demonstrate this with various experiments.

\textbf{The contributions of this paper} can be summarized as follows. We propose a novel disentangled representation learning scheme for \emph{de novo} face image generation via a imitative-contrastive paradigm leveraging 3D priors. Our method enables precise control of the targeted face properties such as pose, expression, and illumination, achieving flexible and high-quality face image generation that, to our knowledge, cannot be achieved by any previous method. 
Moreover, we offer several analyses to understand the properties of the disentangled StyleGAN latent space.
At last, we demonstrate that our method can be used to project real images into the disentangled latent space for analysis and decomposition.


\section{Related Work}
We briefly review the literature on disentangled representation learning and face image synthesis as follows.

\vspace{4pt}
\noindent\textbf{Disentangled representation learning.~} Disentangled representation learning (DRL) for face images has been vividly studied in the past. Historical attempts are based on simple bilinear models~\cite{tenenbaum2000separating}, restricted Boltzmann machines~\cite{desjardins2012disentangling,reed2014learning}, among others. A seminal GAN research along this direction is InfoGAN~\cite{chen2016infogan}.
However, InfoGAN is known to suffer from training instability~\cite{whitney2016understanding}, and there is no guarantee that each latent variable is semantically meaningful~\cite{pmlr-v97-locatello19a,dai2019diagnosing}. InfoGAN-CR~\cite{lin2019infogan} introduces an additional discriminator to identify the latent code under traversal. SD-GAN~\cite{donahue2017semantically} applies a discriminator on image pairs to disentangle identity and appearance factors. Very recently, HoloGAN~\cite{nguyen2019hologan} disentangles 3D pose and identity with unsupervised learning using 3D convolutions and rigid feature transformations. 
DRL with VAEs also received much attention in recent years~\cite{kulkarni2015deep, whitney2016understanding,higgins2017beta,chen2018isolating,kim2018disentangling}.

\vspace{4pt}
\noindent\textbf{Conditional GAN for face synthesis.~} CGAN~\cite{mirza2014conditional} has been widely used in face image synthesis tasks especially identity-preserving generation~\cite{tran2017disentangled,bao2017cvae,yin2017towards,bao2018towards,shen2018faceid}. In a typical CGAN framework, the input to a generator consists of random noises together with some preset conditional factors (\eg, categorical labels or features) as constraints, and an auxiliary classifier/feature extractor is applied to restore the conditional factors from generator outputs. It does not offer a generative modeling of the conditional factors. Later we show that our method can be applied to various face generation tasks handled previously with CGAN frameworks.

\begin{figure*}[t!]
	\small
	\centering
	\includegraphics[width=0.95\textwidth]{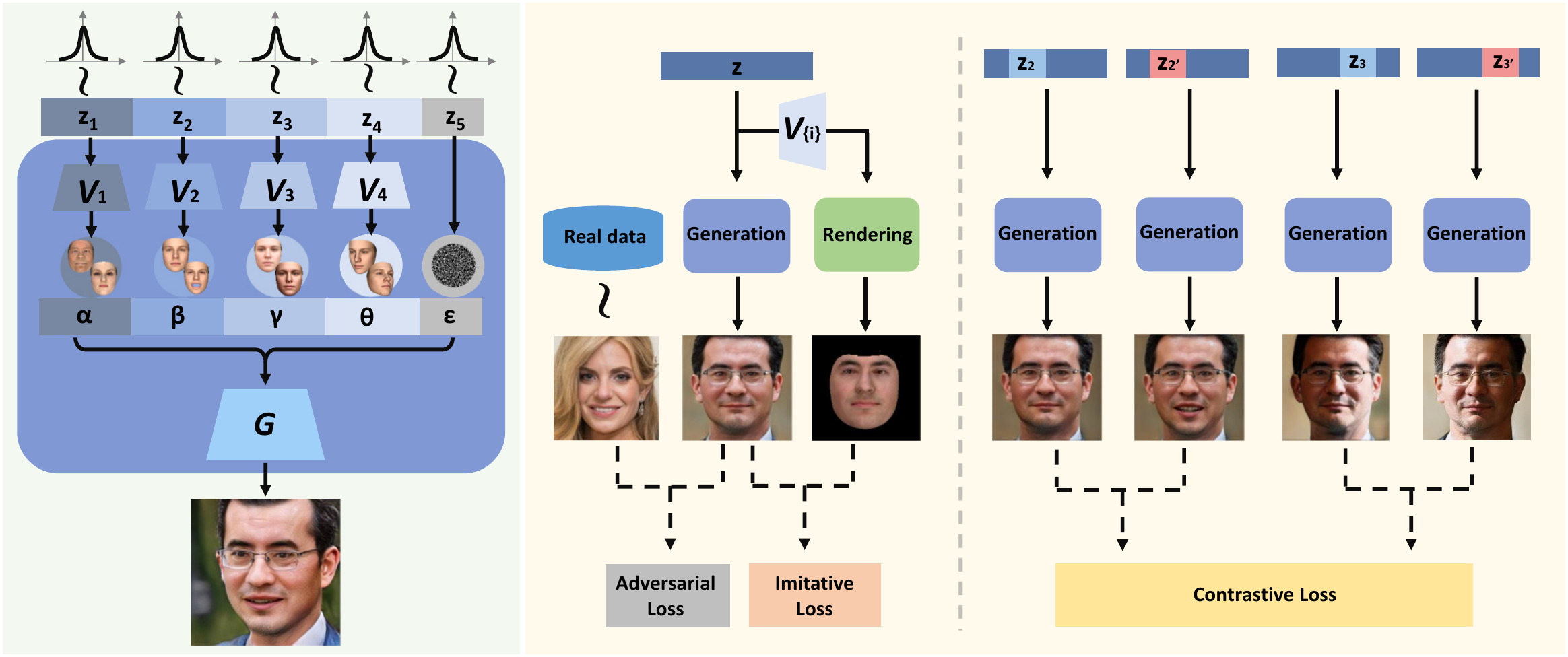}
	\vspace{-8pt}
	\caption{Overview of the \textit{\textbf{DiscoFaceGAN}}. The left diagram (in green) shows the generation pipeline, and the rest illustrates our training scheme which features three type of losses: adversarial loss, imitative loss, and contrastive loss. \label{fig:framework}}
	\vspace{-4pt}
\end{figure*}

\vspace{3pt}
\noindent\textbf{Face image embedding and editing with GANs.~} GANs have seen heavy use in face image manipulation~\cite{perarnau2016icgan,hu2018disentangling,xiao2018elegant,song2018geometry,pumarola2018ganimation,sun2019single,zhou2019deep}. These methods typically share an encoder-decoder/generator-discriminator paradigm where the encoder embeds images into disentangled latent representations characterizing different facial properties. 
Our method can also be applied to embed face images into our disentangled latent space, as we will show in the experiments.

\vspace{4pt}
\noindent\textbf{3D prior for GANs.~} Many methods have been proposed to incorporate 3D prior into GAN for face synthesis~\cite{yin2017towards,shen2018facefeat,kim2018deep,deng2018uv,gecer2018semi,piao2019semi,geng20193d,nguyen2019hologan,xu2020portrait}. Most of them leverages 3DMMs. For example, \cite{kim2018deep} utilizes 3DMM coefficients extracted from input images as low-frequency feature for frontal face synthesis. \cite{gecer2018semi} and \cite{piao2019semi} translate rendered 3DMM faces and real face images in a cycle fashion. \cite{kim2018deep} generates video frames from 3DMM faces for face re-animation. \cite{xu2020portrait} uses 3DMM for portrait reconstruction and pose manipulation. Different from these methods, we only employ 3DMM as priors in the training stage for our imitative-contrastive learning. After training, we do not require a 3DMM model or any rendering procedure.

\section{Approach}

Given a collection of real face images $\mathcal{Y}$, our goal is to train a network $G$ that generates realistic face images $x$ from random noise $z$, which consists of multiple independent variables $z_i\in\mathbb{R}^{N_i}$, each following the normal distribution. We consider latent variables for five independent factors: identity, expression, illumination, pose, and a random noise accounting for other properties such as background. As in standard GAN, a discriminator $D$ is applied to compete with $G$. To obtain disentangled and interpretable latent space, we incorporate 3D priors in an imitative-contrastive learning scheme (Fig. \ref{fig:framework}), described as follows.

\subsection{Imitative Learning}\label{sec:imitative}

To learn how a face image should be generated following the desired properties, we incorporate a 3DMM model~\cite{paysan20093d} and train the generator to imitate the rendered 3D faces. With a 3DMM, the 3D shape $\bS$ and texture $\bT$ of a face is parameterized as 

\vspace{-11pt}

\begin{equation}
	\begin{split}
		\bS & = \bar{\bS} + \bB_{id}\alpha_s + \bB_{exp}\beta \\
		\bT & = \bar{\bT} + \bB_t\alpha_t
	\end{split}\label{equation:MM}
	\vspace{-15pt}
\end{equation}
where $\bar{\bS}$ and $\bar{\bT}$ are the average face shape and texture, $\bB_{id}$,  $\bB_{exp}$, and $\bB_t$ are the PCA bases of identity, expression, and texture, respectively, and $\alpha_s$, $\beta$, and $\alpha_t$ are the corresponding 3DMM coefficient vectors. We denote $\alpha\doteq[\alpha_s,\alpha_t]$ as the identity-bearing coefficients. We approximate scene illumination with Spherical Harmonics (SH)~\cite{ramamoorthi2001efficient} parameterized by coefficient vector $\gamma$. Face pose is defined as three rotation angles\footnote{We align the images to cancel translation.} expressed as vector $\theta$.  With $\lambda \doteq[\alpha, \beta,\gamma,\theta]$, we can easily obtain a rendered face $\hat{x}$ through a well-established analytic image formation~\cite{blanz1999morphable}.

To enable imitation, we first bridge the $z$-space to the $\lambda$-space. We achieve this by training VAE models on the $\lambda$ samples extracted from real image set $\mathcal{Y}$. More specifically, we use the 3D face reconstruction network from \cite{deng2019accurate} to obtain the coefficients of all training images and train four simple VAEs for $\alpha$, $\beta$, $\gamma$ and $\theta$, respectively. After training, we discard the VAE encoders and keep the decoders, denoted as $V_i$, $i\!=\!1,2,3,4$, for $z$-space to $\lambda$-space mapping.

In our GAN training, we sample $\mathrm{z}=[z_1,\ldots,z_5]$ from standard normal distribution, map it to $\lambda$, and feed $\lambda$ to both the generator $G$ and the renderer to obtain a generated face $x$ and a rendered face $\hat{x}$, respectively. Note that we can input either $\mathrm{z}$ or $\lambda$ into $G$ -- in practice we observe no difference between these two options in terms of either visual quality or disentangling efficacy. The benefit of using $\lambda$ is the ease of face property control since $\lambda$ is interpretable.

We define the following loss functions on $x$ for imitative learning. First, we enforce $x$ to mimic the identity of $\hat{x}$ perceptually by 

\vspace{-7pt}

\begin{equation}
	l^{id}_{I}(x) = \max(1-<f_{id}(x),f_{id}(\hat{x})>-\tau, 0),
	\vspace{0pt}
\end{equation}
where $f_{id}(\cdot)$ is the deep identity feature from a face recognition network, $<\!\cdot,\cdot\!>$ denotes cosine similarity, and $\tau$ is a constant margin which we empirically set as $0.3$. Since there is an obvious domain gap between rendered 3DMM faces and real ones, we allow a small difference between the features. The face recognition network from \cite{yang2017neural} is used in this paper for deep identity feature extraction. For expression and pose, we penalize facial landmark differences via

\vspace{-8pt}

\begin{equation}{}\label{loss_lm}
	l^{lm}_{I}(x) = \|p(x)-\hat{p}\|^2,
\end{equation}
where $p(\cdot)$ denotes the landmark positions detected by the 3D face reconstruction network, and $\hat{p}$ is the landmarks of the rendered face obtained trivially. For illumination, we simply minimize the SH coefficient discrepancy by
\begin{equation}
	l^{sh}_{I}(x) = |\gamma(x)-\hat{\gamma}|_1,
\end{equation}
where $\gamma(\cdot)$ represents the coefficient given by the 3D face reconstruction network, and $\hat{\gamma}$ is the coefficient of $\hat{x}$. 
Finally, we add a simple loss which enforces the output to mimic the skin color of the rendered face via 
\begin{equation}
	l^{cl}_{I}(x) = |c(x)-c(\hat{x})|_1,
\end{equation}
where $c(\cdot)$ denotes the average color of face region defined by the mask in 3DMM. By using these imitative losses, the generator will learn to generate face images following the identity, expression, pose, and illumination characterized by the corresponding latent variables.

\vspace{5pt}
\noindent\textbf{The domain gap issue.~} Obviously, there is an inevitable domain gap between the rendered 3DMM faces and generated ones. Understanding the effect of this domain gap and judiciously dealing with it is important. On one hand, retaining a legitimate domain gap that is reasonably large is necessary as it avoids the conflict with the adversarial loss and ensures the realism of generated images. It also prevents the generative modeling from being trapped into the small identity subspace of the 3DMM model\footnote{The 3DMM we use in this paper is from \cite{paysan20093d} which is constructed by scans of 200 people.}.
On the other hand, however, it may lead to poor factor variation disentanglement (for example, changing expression may lead to unwanted variations of identity and image background, and changing illumination may disturb expression and hair structure; see Fig.~\ref{fig:warp} and \ref{fig:ablation}).

To understand why this happens, we first symbolize the difference between a generated face $x$ and its rendered counterpart $\hat{x}$ as $\Delta x$, \ie, $x = \hat{x}  + \Delta x$. In the imitative learning, $x$ is free to deviate from $\hat{x}$ in terms of certain identity characteristics and other image contents beyond face region (\eg, background, hair, and eyewear). As a consequence, $\Delta x$ has a certain degree of freedom that is \emph{uncontrollable}. We resolve this issue via contrastive learning, to be introduced next. 

\subsection{Contrastive Learning}

To fortify disentanglement, we enforce the invariance of the latent representations for image generation in a contrastive manner: we vary one latent variable while keeping others unchanged, and enforce that the difference on the generated face images relates only to that latent variable. Concretely, we sample pairs of latent code $\mathrm{z}, \mathrm{z}'$ which differ only at $z_k$ and share the same $z_i, \forall i\neq k$. We compare the generated face images $x$, $x'$, and then penalize the difference induced by any of $z_i$ but $z_k$. 

\begin{figure}[t!]
	\small
	\centering
	\includegraphics[width=\columnwidth]{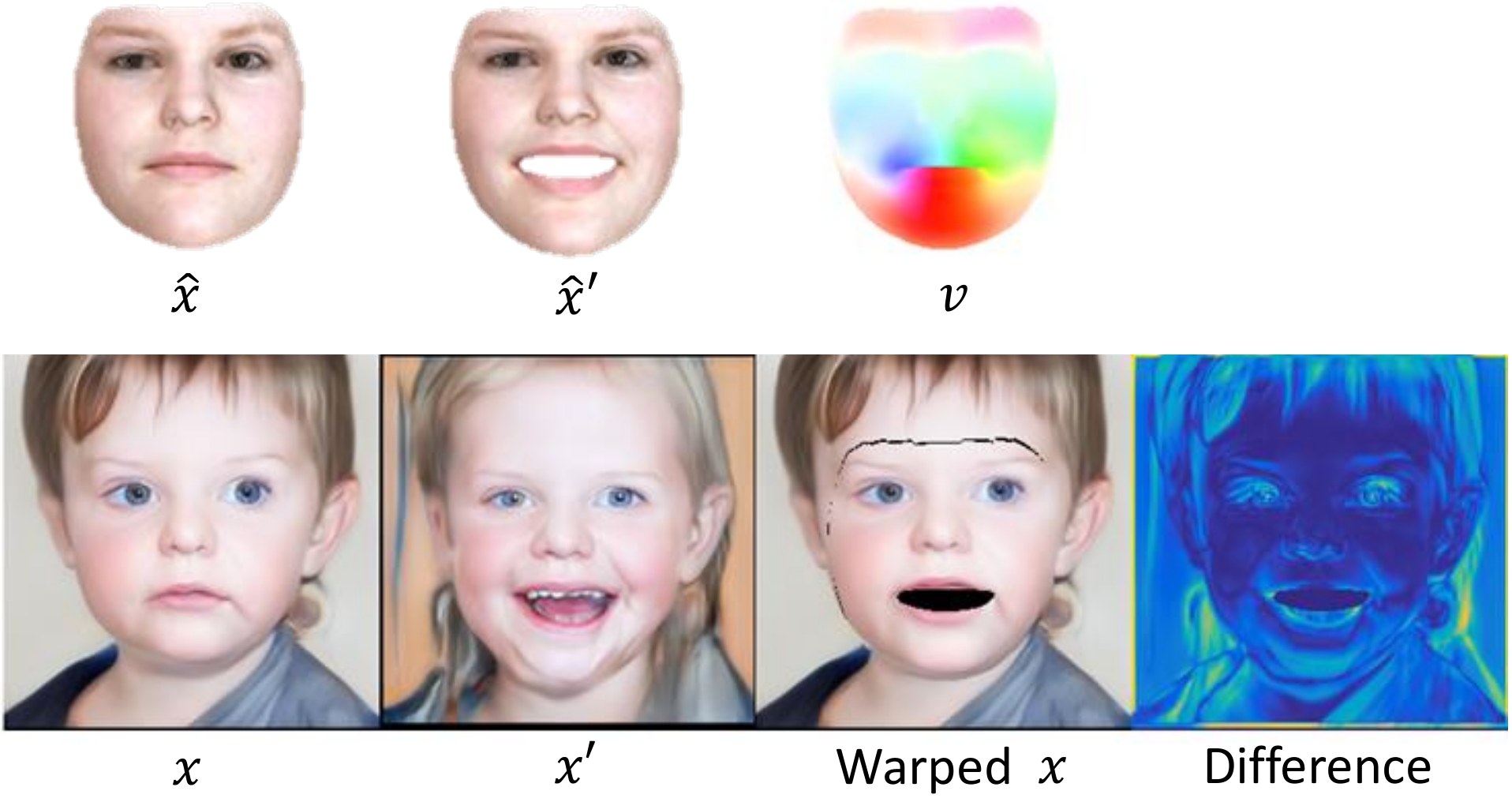}
	\vspace{-20pt}
	\caption{Illustration of the image warping process in our contrastive learning.\label{fig:warp}}
	\vspace{-4pt}
\end{figure}

To enable such a comparison, we need to find a function $\phi_k(G(\mathrm{z}))$ which is, to the extent possible, invariant to $z_k$ but sensitive to variations of $z_i$'s. In this work, we implement two simple functions for face images. The first one is designed for expression-invariant comparison.
Our idea is to restore a neutral expression for $x$ and $x'$ to enable the comparison. However,  high-fidelity expression removal \emph{per se} is a challenging problem still being actively studied in GAN-based face image manipulation~\cite{qian2019unsupervised,geng20193d}. To circumvent this issue, we resort to the rendered 3DMM face $\hat{x}$ to get a surrogate flow field for image warping. Such a flow field can be trivially obtained by revising the expression coefficient and rendering another 3DMM face with a neutral expression. In practice, it is unnecessary to warp both $x$ and $x'$. We simply generate the flow field $v$ from $\hat{x}$ to $\hat{x}'$ and warp $x$ to $x'$ accordingly (see Fig.~\ref{fig:warp} for an example). We then minimize the image color difference via

\vspace{-5pt}

\begin{equation}
	l^{ex}_{C}(x,x') = |x(v)-x'|_1,
\end{equation}
where $x(v)$ is the warped image.

Second, we design two illumination-invariant losses for contrastive learning. Since the pixel color across the whole image can be affected by illumination change, we simply enforce the semantical structure to remain static. We achieve this by minimizing the difference between the face structures of $x$ and $x'$:
\vspace{-8pt}

\begin{equation}
	l^{il_1}_{C}(x,x') = \|m(x)-m(x')\|^2 + \omega \|p(x)-p(x')\|^2,
	\vspace{-3pt}
\end{equation}
where $m(\cdot)$ is the hair segmentation probability map obtained from a face parsing network~\cite{lin2019face}, $p(\cdot)$ denotes landmark positions same as in Eq.~\ref{loss_lm}, and $\omega$ is a balancing weight. We also apply a deep identity feature loss via
\begin{equation}
	l^{il_2}_{C}(x,x') = 1- <f_{id}(x), f_{id}(x')>.
\end{equation}

In this paper, using the above contrastive learning losses regarding expression and illumination can lead to satisfactory disentanglement (we found that pose variations can be well disentangled without need for another contrastive loss).

\begin{figure*}[t]
	\includegraphics[width=0.98\textwidth]{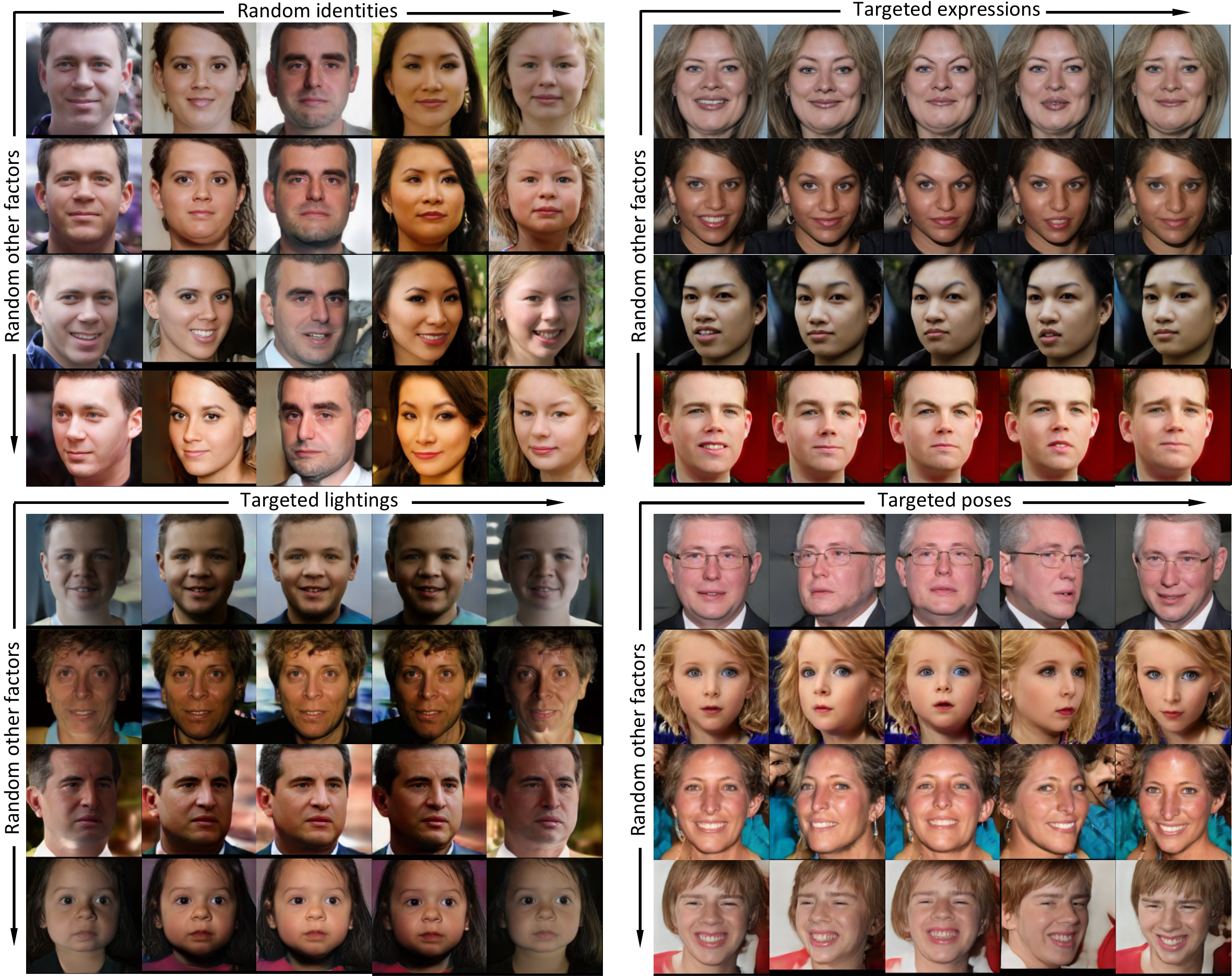}
	\vspace{-5pt}
	\caption{Face images generated by our \textit{DiscoFaceGAN}. As shown in the figures, the variations of identity, expression, pose and illumination are highly disentangled, and we can precisely control expression, illumination and pose.
	}\label{fig:generate_result}
	\vspace{-2pt}
\end{figure*}

\vspace{4pt}
\noindent\textbf{Effect of contrastive learning.~} Following the discussion in Section~\ref{sec:imitative}, for two rendered faces $\hat{x}$ and $\hat{x}'$ which only (and perfectly) differ at one factor such as expression, both $\Delta x$ and $\Delta x'$ have certain free variations that are uncontrollable. Therefore, achieving complete disentanglement with imitative learning is difficult, if not impossible. The contrastive learning is an essential complement to imitative learning: it imposes proper constrains on $\Delta x$ and $\Delta x'$ by explicitly learning the desired differences between $x$ and $x'$, thus leading to enhanced disentanglement.

We empirically find that the contrastive learning also leads to better imitation and more accurate face property control. This is because the pairwise comparison can also suppress imitation noise: any misalignment of pose or expression between $x$ and $\hat{x}$ or between $x'$ and $\hat{x}'$ will incur larger contrastive losses.

\section{Experiments}

\vspace{-2pt}
\noindent\textbf{Implementation details.~}
In this paper, we adopt the StyleGAN structure~\cite{karras2019style} and the FFHQ dataset~\cite{karras2019style} for training.
We train the $\lambda$-space VAEs following the schedule of \cite{dai2019diagnosing}, where encoders and decoders of the VAEs are all MLPs with three hidden layers. For StyleGAN, we follow the standard training procedure of the original method except that we 1) remove the normalization operation for input latent variable layer, 2) discard the style-mixing strategy, and 3) train up to image resolution of $256\times256$ due to time constraint. We first train the
network with the adversarial loss as in~\cite{karras2019style} and our imitative losses until seeing $15M$ real images to obtain reasonable imitation. Then we add contrastive losses into the training process and train the network up to seeing $20M$ real images in total. More training details can be found in the \emph{suppl. material}. 

\subsection{Generation Results}
Figure~\ref{fig:generate_result} presents some image samples generated by our \textit{DiscoFaceGAN} after training. It can be seen that our method is able to randomly generate high-fidelity face images with a large variant of identities with diverse pose, illumination, and facial expression. More importantly, the variations of identity, expression, pose, and illumination are highly disentangled -- when we vary one factor, all others can be well preserved. Furthermore, we can precisely control expression, illumination and pose using the parametric model coefficients for each of them. One more example for precisely controlled generation is given in Fig.~\ref{fig:teaser}.

\begin{figure}[t]
	\centering
	\includegraphics[width=1.0\columnwidth]{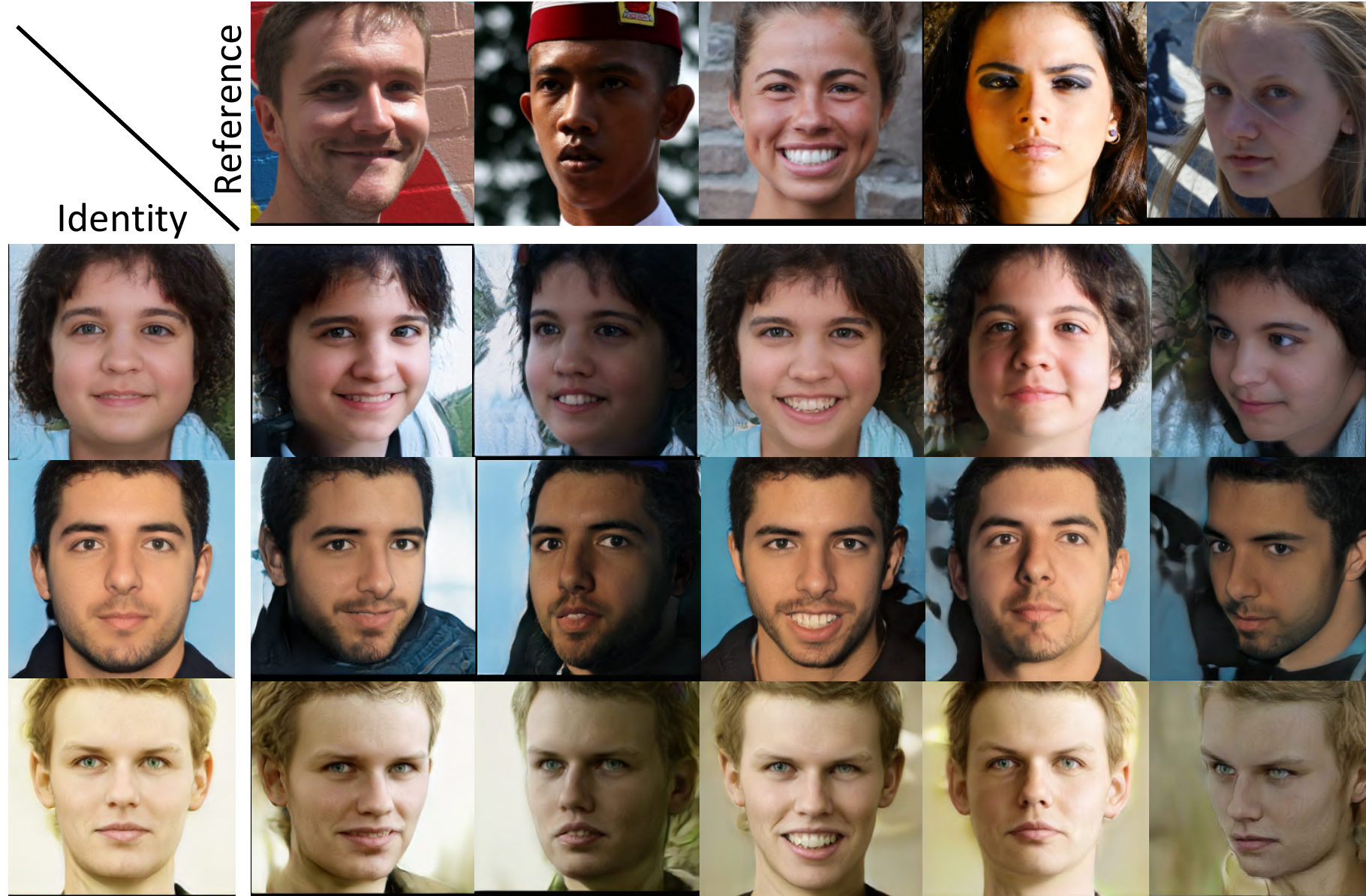}
	\vspace{-18pt}
	\caption{Reference-based generation results where we extract expression, lighting, and pose properties of real images and combine them with randomly generated identities.}\label{fig:refer}
\end{figure}

Figure~\ref{fig:refer} shows that we can generate images of new identities by mimicking the properties of a real reference image. We achieve this by extracting the expression, lighting, and pose parameters from the reference image and combine them with random identity variables for generation.

\subsection{Ablation Study}

\begin{figure}[t]
	\centering
	\includegraphics[width=0.73\columnwidth]{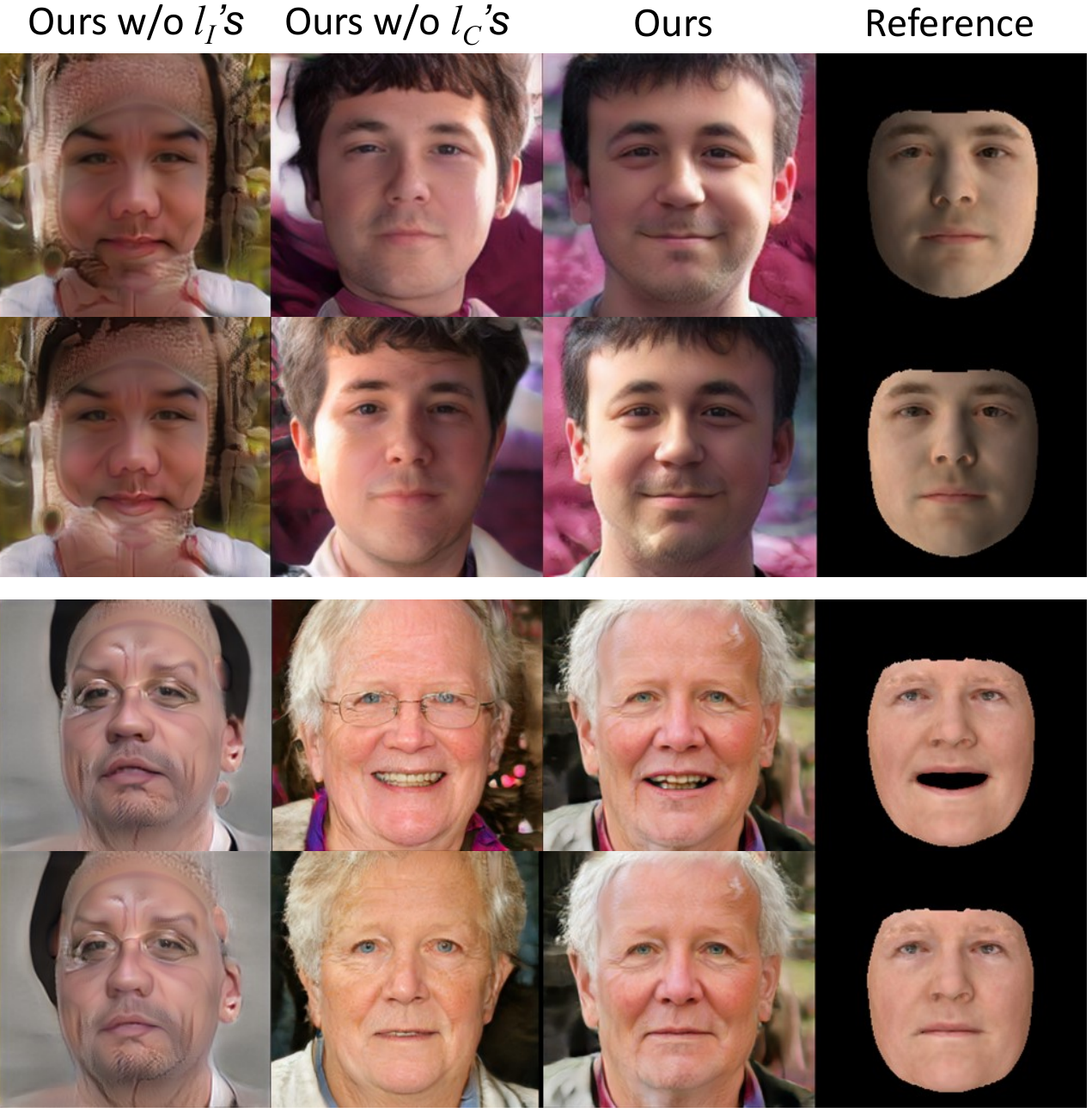}
	\vspace{-7pt}
	\caption{Ablation study of the training losses. The top and bottom two rows show the results when we vary the latent variable for lighting and expression, respectively.
	}\label{fig:ablation}
\end{figure}

In this section, we train the \textit{DiscoFaceGAN} with different losses to validate the effectiveness of our imitative-contrastive learning scheme. Some typical results are presented in Fig.~\ref{fig:ablation}. Obviously, the network cannot generate reasonable face images if we remove the imitation losses. This is because the contrastive losses rely on reasonable imitation, without which they are less meaningful and the network behavior will be unpredictable. On the other hand, without contrastive losses, variations of different factors cannot be fully disentangled. For example, expression and lighting changes may influence certain identity-related characteristics and some other properties such as hair structure. The contrastive losses can also improve the desired preciseness of imitation (\eg, see the mouth-closing status in the last row), leading to more accurate generation control.

\begin{table}[t]
	\centering
	\small
	\caption{Comparison of disentanglement score as well as generation quality.\label{tab:quantity}}
	\vspace{-7pt}
	\begin{tabular}{c|cccc|cc}
		\hline
		& \multicolumn{4}{c|}{Disentanglement $\uparrow$} & \multicolumn{2}{c}{Quality $\downarrow$} \\
		\hline
		\!\!\!\! & \!\!$DS_{{\bf \alpha}}$\!\! & \!\!$DS_{{\bf \beta}}$\!\! & \!\!$DS_{{\bf \gamma}}$\!\!& \!\!$DS_{{\bf \theta}}$\!\! & \!\!FID\!\!&\!\!PPL\!\! \\
		\hline
		\!\!3DMM\!\!& \multicolumn{4}{c|}{-}&271\!&\!-\! \\
		\hline
		\!$l_{adv}$\!& \!0.83\! & \!1.98\! &\! 0.87\! & \!0.07\!& \!\textbf{5.49}\!& \!106\!\\
		\!$+ l_{I}'s$\!& \!\textbf{13.4}\! & \!37.0\! & \!40.4\! & \!31.6& \!9.15\! & \!\textbf{102}\!\\
		\!$+ l_{C}'s$\!& \!7.85\! & \!\!\textbf{80.4}\!\! & \!\textbf{489}\! & \!\textbf{36.7}\!& \!12.9\! &\!123\!\\
		\hline
	\end{tabular}
\end{table}

\subsection{Quantitative Evaluation}

In this section, we evaluate the performance of our \textit{DiscoFaceGAN} quantitatively in terms of disentanglement efficacy as well as generation quality. For the former, several metrics have been proposed in VAE-based disentangled representation learning, such as factor score~\cite{kim2018disentangling} and mutual information gap~\cite{chen2018isolating}. However, these metrics are not suitable for our case. Here we design a simple metric named disentanglement score (DS), described as follows.

Our goal is to measure that when we only vary the latent variable for one single factor, if other factors on the generated images are stable.
We denote the four $\lambda$-space variables ${\bf \alpha},{\bf \beta},{\bf \gamma},{\bf \theta}$ as $\tmp_i$, and we use $\tmp_{\{j\}}$ as the shorthand notation for the variable set $\{\tmp_{j}|j=1,\ldots,4, j\neq i\}$.  
To measure the disentanglement score for $u_i$, we first randomly generate 1K sets of $\tmp_{\{j\}}$, and for each $\tmp_{\{j\}}$ we randomly generate 10 $u_i$. Therefore, we can generate 10K images using the trained network with combinations of $u_i$ and $\tmp_{\{j\}}$. For these images, we re-estimate $u_i$ and $\tmp_{\{j\}}$ using the 3D reconstruction network~\cite{deng2019accurate} (for identity we use a face recognition network~\cite{yang2017neural} to extract deep identity feature instead). We calculate the variance of the estimated values for each of the 1K groups, and then average them to obtain $\sigma_{u_i}$ and $\sigma_{u_j}$. We further normalize $\sigma_{u_i}$ and $\sigma_{u_j}$ by dividing the variance of the corresponding variable computed on FFHQ. Finally, we measure the disentanglement score via 
\begin{equation}
DS_{u_i} = \Pi_{j,j\neq i}\frac{\sigma_{u_i}}{\sigma_{u_j}},
\label{eq:dcs}
\end{equation}A high $DS$ indicates that when varying a certain factor, only the corresponding property in the generated images is changing ($\sigma_{u_i}>0$) while other factors remain unchanged ($\sigma_{u_j}\rightarrow 0$). Table \ref{tab:quantity} shows that the imitative learning  leads to high factor disentanglement and the contrastive learning further enhances it for expression, illumination, and pose. The disentanglement score for identity decreases with contrastive learning. We found the 3D reconstruction results from the network are slightly unstable when identity changes, which increased the variances of other factors.

To evaluate the quality of image generation, we follow \cite{karras2019style} to compute the Fr\'echet Inception Distances (FID)~\cite{heusel2017gans} and the Perceptual Path Lengths (PPL)~\cite{karras2019style} using 50K and 100K randomly generated images, respectively.
Table \ref{tab:quantity} shows that the FID increases with our method. This is expected as the additional losses added to the adversarial training will inevitably affect the generative modeling. However, we found that the PPL is comparable to the results trained with only the adversarial loss.

\section{Latent Space Analysis and Embedding}

In this section, we analyze the latent space of our \textit{DiscoFaceGAN}. We show some meaningful properties supporting factor variation disentanglement, based on which we further present a method for embedding and manipulating real face images in the disentangled latent space.
\subsection{Analysis of Latent Space}\label{sec:latent_space}
One key ingredients of StyleGAN is the mapping from $z$-space to $\mathcal{W}$-space, the latter of which relates linearly to the AdaIN~\cite{huang2017arbitrary} parameters that control ``styles" (we refer the readers to \cite{karras2019style} for more details). Previous studies~\cite{shen2019interpreting,abdal2019image2stylegan} have shown that certain \emph{direction of changes} in $\mathcal{W}$-space leads to variations of corresponding attributes in generated images.
In our case, $\mathcal{W}$ space is mapped from $\lambda$ space which naturally relates to image attributes. Therefore, we analyze the \emph{direction of changes} in the learned $\mathcal{W}$-space by varying $\lambda$ variables, and some interesting properties have been found. We will introduce these properties and then provide strong empirical evidences supporting them. 

Recall that the input to generator is $\lambda$-space variables ${\bf \alpha},{\bf \beta},{\bf \gamma},{\bf \theta}$ and an additional noise $\varepsilon$. Here we denote these five variables as $\tmp_i$ with $\tmp_5=\varepsilon$. We use $\tmp_{\{j\}}$ as the shorthand notation for the variable set $\{\tmp_{j}|j=1,\ldots,5, j\neq i\}$, and $w(\tmp_i, \tmp_{\{j\}})$ denotes the $\mathcal{W}$ space variable mapped from $\tmp_i$ and $\tmp_{\{j\}}$. We further denote a unit vector
\begin{equation}
\widehat{\Delta w}(i,a,b) = \frac{w(\tmp_i\!=\!a, \tmp_{\{j\}}\!\large)-w\large(\tmp_i\!=\!b, \tmp_{\{j\}}\!)}{\|w(\tmp_i\!=\!a, \tmp_{\{j\}}\!) -w(\tmp_i\!=\!b, \tmp_{\{j\}}\!)\|}\label{eq:deltaw}
\end{equation}
to represent the direction of change in $\mathcal{W}$ space when we change $u_i$ from $a$ to $b$. The following two properties of $\widehat{\Delta w}(i,a,b)$ are observed:

\vspace{4pt}
\noindent\textbf{Property 1.~}
For the $i$-th variable $\tmp_i$, $i\in{1,2,3,4}$, with any given starting value $a$ and ending value $b$, we have:
\vspace{-7pt}
\begin{equation}
\begin{split}
\widehat{\Delta w}(i,a,b)~\text{\emph{is almost constant for}~} \forall \tmp_{\{j\}}\text{\emph{.}}
\\
\end{split}\label{eq:property1} \nonumber
\end{equation}

\vspace{2pt}
\noindent\textbf{Property 2.~}
For the $i$-th variable $\tmp_i$, $i\in{1,2,3,4}$, with any given offset vector $\triangle$, we have:
\vspace{-7pt}
\begin{equation}
\begin{split}
\widehat{\Delta w}(i,a,a\!+\!\triangle)~\text{\emph{is almost constant for}~} \forall \tmp_{\{j\}}\text{\emph{and}~} \forall a{\emph{.}}\\
\end{split}\label{eq:property2} \nonumber
\end{equation}
\vspace{-15pt}

\noindent Property~1 states that if the starting and ending values of a certain factor in $\lambda$ space are fixed, then the direction of change in $\mathcal{W}$ space is stable regardless of the choice of all other factors. Property~2 further indicates that it is unnecessary to fix the starting and ending values -- the direction of change in $\mathcal{W}$ space is only decided by the difference between them.

\begin{table}[t]
	\centering
	\small
	\caption{Cosine similarities of direction of change in $\mathcal{W}$ space. \textbf{Top:} changing a factor from a fixed start to a fixed end. \textbf{Bottom:} changing a factor with a fixed offset.\label{tab:properties}}
	\vspace{-7pt}
	\begin{tabular}{ccccc}
		\hline
		& \!identity\! & \!expression\! & \!lighting\! & \!pose\!\\
		\hline
		\!$l_{adv}$\!& \!0.65$\pm$ 0.10\! & \!0.21$\pm$ 0.11\! &\! 0.16$\pm$ 0.12\! & \!0.17$\pm$ 0.11\! \\
		\!Ours\!& \!\textbf{0.96$\pm$ 0.02}\! & \!\textbf{0.82$\pm$ 0.04}\! & \!\textbf{0.85$\pm$ 0.03}\! & \!\textbf{0.87$\pm$ 0.03}\! \\
		\hline
	\end{tabular}
	\vspace{5pt}.
	\begin{tabular}{ccccc}
		\hline
		& \!identity\! & \!expression\! & \!lighting\! & \!pose\!\\
		\hline
		\!$l_{adv}$\!& \!0.42$\pm$ 0.14\! & \!0.21$\pm$ 0.12\! &\! 0.16$\pm$ 0.12\! & \!0.15$\pm$ 0.11\! \\
		\!Ours\!& \!\textbf{0.82$\pm$ 0.06}\! & \!\textbf{0.79$\pm$ 0.05}\! & \!\textbf{0.85$\pm$ 0.04}\! & \!\textbf{0.85$\pm$ 0.04}\! \\
		\hline
	\end{tabular}
\end{table}

To empirically examine Property~1, we randomly sampled 50 pairs of $(a,b)$ values for each $\tmp_i$ and 100 remaining factors for each pair. For each $(a,b)$ pair, we calculate 100 $\Delta w = w_2 - w_1$ and get $100\times 100$ pairwise cosine distances. We average all these distances for each $(a,b)$ pair, and finally compute the mean and standard derivation of the 50 average distance values from all 50 pairs.
Similarly, we examine Property~2 by randomly generating offsets for $\tmp_i$, and all the results are presented in Table~\ref{tab:properties}. It can be seen that all the cosine similarities are close to 1, indicating the high consistency of $\mathcal{W}$-space direction change.
For reference, in the table we also present the statistics obtained using a model trained with the same pipeline but without our imitative-contrastive losses.

\subsection{Real Image Embedding and Editing}\label{sec:manipulation}

Based on the above analysis, we show that our method can be used to embed real images into the latent space and edit the factors in a disentangled manner. We present the experimental results on various factors. More results can be found in the \emph{suppl. material} due to space limitation.

A natural latent space for image embedding and editing is the $\lambda$ space.
However, embedding an image to it leads to poor image reconstruction.
Even inverting to the $\mathcal{W}$ space is problematic -- the image details are lost as shown in previous works~\cite{abdal2019image2stylegan,shen2019interpreting}.
For higher fidelity, we embed the image into a latent code $w^+$ in the $\mathcal{W}+$ space suggested by \cite{abdal2019image2stylegan} which is an extended $\mathcal{W}$ space. An optimization-based embedding method is used similar to \cite{abdal2019image2stylegan}.
However, $\mathcal{W}$ or $\mathcal{W+}$ space is not geometrically interpretable thus cannot be directly used for controllable generation.
Fortunately though, thanks to the nice properties of the learned $\mathcal{W}$ space (see Section~\ref{sec:latent_space}), 
we have the following latent representation editing and image generation method:
\begin{equation}
\begin{split}
w^+_{s} =&~ G_{syn}^{-1}(x_{s})\\
x_{t} =&~ G_{syn}(w^+_{s} + \Delta w(i,a,b))\\
\end{split}\label{equation:editing}
\end{equation}
where $x_s$ is an input image and $x_t$ is the targeted image after editing. $G_{syn}$ is the synthesis sub-network of StyleGAN (after the 8-layer MLP). $\Delta w(i,a,b)$ denotes the offset of $w$ induced by changing $u_i$, the $i$-th $\lambda$-space latent variable, from $a$ to $b$ (see Eq.~\ref{eq:deltaw}). It can be computed with any $\tmp_{\{j\}}$ (we simply use the embedded one). Editing can be achieved by flexibly setting $a$ and $b$.

\begin{figure}[t!]
	\includegraphics[width=0.982\columnwidth]{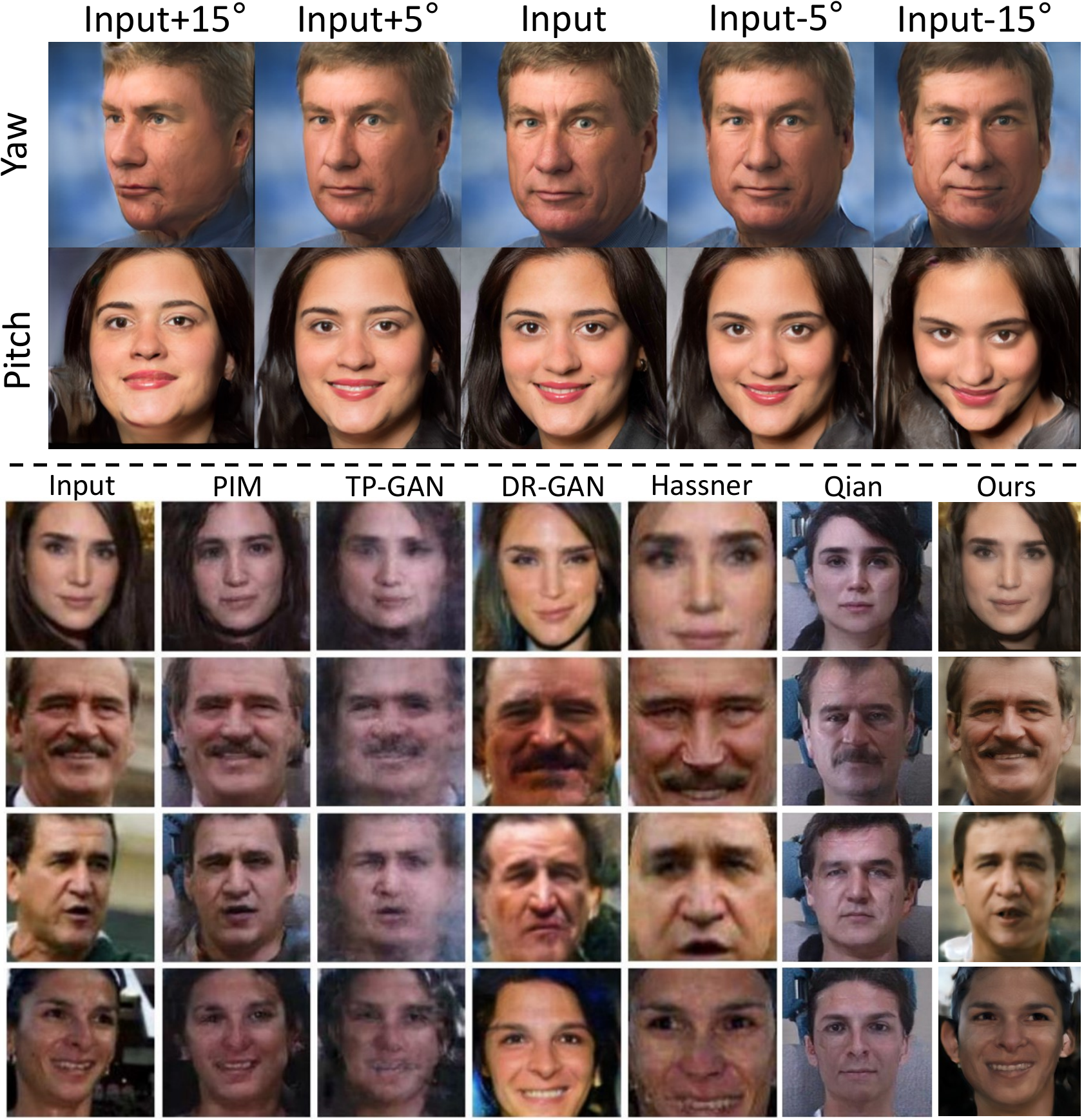}
	\vspace{-6pt}
	\caption{Real image pose manipulation results. \textbf{Top}: Precise manipulation of pose angles. \textbf{Bottom}: Face frontalization results compared with PIM \cite{zhao2018towards}, TP-GAN \cite{huang2017beyond}, DR-GAN \cite{tran2017disentangled}, Hassner \etal~\cite{hassner2015effective}, and Qian \etal~\cite{qian2019unsupervised} on LFW. Results of other methods are from \cite{qian2019unsupervised}.}\label{fig:pose}
	\vspace{-2pt}
\end{figure}

\vspace{5pt}
\noindent\textbf{Pose Editing.~}
Figure~\ref{fig:pose} (top) shows the typical results of pose manipulation where we freely rotate the input face by desired angles.
We also test our method with the task of face frontalization, and compare with previous methods.
Figure~\ref{fig:pose} (bottom) shows the results on face images from the LFW dataset~\cite{huang2008labeled}. Our method well-preserved the identity-bearing characteristics as well as other contextual information such as hair structure and illumination.

\begin{figure}[t!]
	\includegraphics[width=0.99\columnwidth]{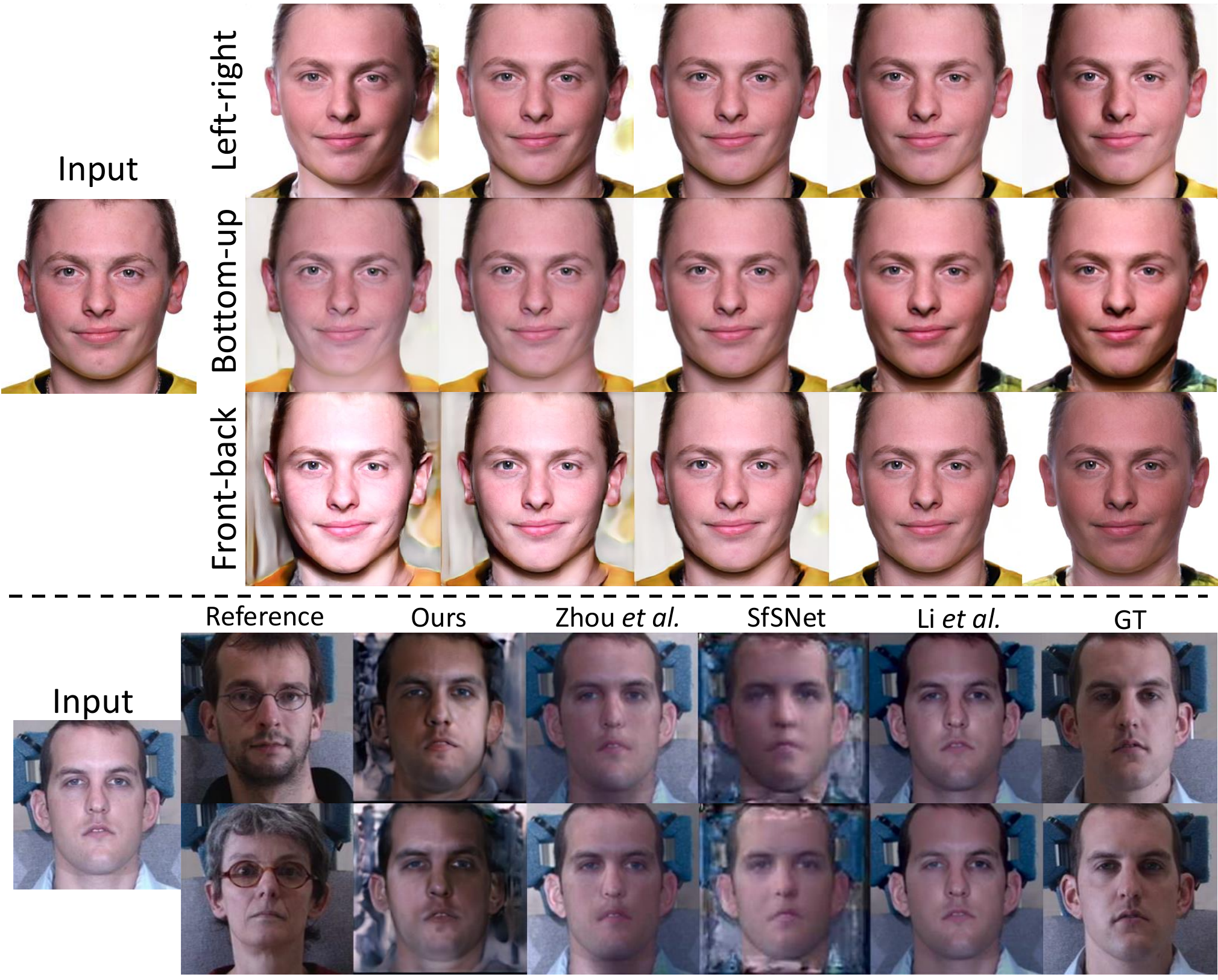}
	\vspace{-8pt}
	\caption{Real image relighting results. \textbf{Top}: Light editing for a real image. \textbf{Bottom}: Results on a challenging lighting transfer task compared with Zhou \etal~\cite{zhou2019deep}, SfSNet~\cite{sengupta2018sfsnet}, and Li \etal~\cite{li2018closed}. Results of other methods are from \cite{zhou2019deep}.}\label{fig:relight}
	\vspace{-5pt}
\end{figure}

\vspace{5pt}
\noindent\textbf{Image Relighting.~}
Figure~\ref{fig:relight} (top) shows an example of image relighting with our method, where we freely vary the lighting direction and intensity. In addition, we follow the previous methods to evaluate our method on the MultiPIE~\cite{gross2010multi} images. Figure~\ref{fig:relight} (bottom) shows a challenging case for lighting transfer.
Despite the extreme indoor lighting may be outside of the training data, our method still produces reasonable results with lighting directions well conforming with the references.

\section{Conclusion and Future Work}
We presented \textit{\textbf{DiscoFaceGAN}} for disentangled and controllable latent representations for face image generation. The core idea is to incorporate 3D priors into the adversarial learning framework and train the network to imitate the rendered 3D faces. Influence of the domain gap between rendered faces and real images is properly handled by introducing the contrastive losses which explicitly enforce disentanglement. Extensive experiments on disentangled virtual face image synthesis and face image embedding have demonstrated the efficacy of our proposed imitation-contrastive learning scheme.

The generated virtual identity face images with accurately controlled properties could be used for a wide range of vision and graphics applications which we will explore in our future work. It is also possible to apply our method for forgery image detection and anti-spoofing by analyzing real and faked images in the disentangled space.

{\small
	\bibliographystyle{ieee_fullname}
	\bibliography{facesynthesis}

\begin{thebibliography}{10}\itemsep=-1pt

\bibitem{abdal2019image2stylegan}
Rameen Abdal, Yipeng Qin, and Peter Wonka.
\newblock Image2stylegan: How to embed images into the stylegan latent space?
\newblock In {\em IEEE International Conference on Computer Vision}, pages
  4432--4441, 2019.

\bibitem{bao2017cvae}
Jianmin Bao, Dong Chen, Fang Wen, Houqiang Li, and Gang Hua.
\newblock {CVAE-GAN}: fine-grained image generation through asymmetric
  training.
\newblock In {\em IEEE International Conference on Computer Vision}, pages
  2745--2754, 2017.

\bibitem{bao2018towards}
Jianmin Bao, Dong Chen, Fang Wen, Houqiang Li, and Gang Hua.
\newblock Towards open-set identity preserving face synthesis.
\newblock In {\em IEEE Conference on Computer Vision and Pattern Recognition},
  pages 6713--6722, 2018.

\bibitem{blanz1999morphable}
Volker Blanz, Thomas Vetter, et~al.
\newblock A morphable model for the synthesis of 3d faces.
\newblock In {\em SIGGRAPH}, volume~99, pages 187--194, 1999.

\bibitem{chen2018isolating}
Tian~Qi Chen, Xuechen Li, Roger~B Grosse, and David~K Duvenaud.
\newblock Isolating sources of disentanglement in variational autoencoders.
\newblock In {\em Advances in Neural Information Processing Systems}, pages
  2610--2620, 2018.

\bibitem{chen2016infogan}
Xi Chen, Yan Duan, Rein Houthooft, John Schulman, Ilya Sutskever, and Pieter
  Abbeel.
\newblock {InfoGAN}: Interpretable representation learning by information
  maximizing generative adversarial nets.
\newblock In {\em Advances in Neural Information Processing Systems}, pages
  2172--2180, 2016.

\bibitem{dai2019diagnosing}
Bin Dai and David Wipf.
\newblock Diagnosing and enhancing vae models.
\newblock {\em arXiv preprint arXiv:1903.05789}, 2019.

\bibitem{deng2018uv}
Jiankang Deng, Shiyang Cheng, Niannan Xue, Yuxiang Zhou, and Stefanos
  Zafeiriou.
\newblock {UV-GAN}: Adversarial facial uv map completion for pose-invariant
  face recognition.
\newblock In {\em IEEE Conference on Computer Vision and Pattern Recognition},
  pages 7093--7102, 2018.

\bibitem{deng2019accurate}
Yu Deng, Jiaolong Yang, Sicheng Xu, Dong Chen, Yunde Jia, and Xin Tong.
\newblock Accurate 3d face reconstruction with weakly-supervised learning: From
  single image to image set.
\newblock In {\em IEEE Computer Vision and Pattern Recognition Workshop on
  Analysis and Modeling of Faces and Gestures}, 2019.

\bibitem{desjardins2012disentangling}
Guillaume Desjardins, Aaron Courville, and Yoshua Bengio.
\newblock Disentangling factors of variation via generative entangling.
\newblock {\em arXiv preprint arXiv:1210.5474}, 2012.

\bibitem{donahue2017semantically}
Chris Donahue, Zachary~C Lipton, Akshay Balsubramani, and Julian McAuley.
\newblock Semantically decomposing the latent spaces of generative adversarial
  networks.
\newblock In {\em International Conference on Learning Representations}, 2018.

\bibitem{gecer2018semi}
Baris Gecer, Binod Bhattarai, Josef Kittler, and Tae-Kyun Kim.
\newblock Semi-supervised adversarial learning to generate photorealistic face
  images of new identities from 3d morphable model.
\newblock In {\em European Conference on Computer Vision}, pages 217--234,
  2018.

\bibitem{geng20193d}
Zhenglin Geng, Chen Cao, and Sergey Tulyakov.
\newblock 3d guided fine-grained face manipulation.
\newblock In {\em IEEE Conference on Computer Vision and Pattern Recognition},
  pages 9821--9830, 2019.

\bibitem{goodfellow2014generative}
Ian Goodfellow, Jean Pouget-Abadie, Mehdi Mirza, Bing Xu, David Warde-Farley,
  Sherjil Ozair, Aaron Courville, and Yoshua Bengio.
\newblock Generative adversarial nets.
\newblock In {\em Advances in Neural Information Processing Systems}, pages
  2672--2680, 2014.

\bibitem{gross2010multi}
Ralph Gross, Iain Matthews, Jeffrey Cohn, Takeo Kanade, and Simon Baker.
\newblock Multi-pie.
\newblock {\em Image and Vision Computing}, pages 807--813, 2010.

\bibitem{hassner2015effective}
Tal Hassner, Shai Harel, Eran Paz, and Roee Enbar.
\newblock Effective face frontalization in unconstrained images.
\newblock In {\em IEEE Conference on Computer Vision and Pattern Recognition},
  pages 4295--4304, 2015.

\bibitem{heusel2017gans}
Martin Heusel, Hubert Ramsauer, Thomas Unterthiner, Bernhard Nessler, and Sepp
  Hochreiter.
\newblock Gans trained by a two time-scale update rule converge to a local nash
  equilibrium.
\newblock In {\em Advances in neural information processing systems}, pages
  6626--6637, 2017.

\bibitem{higgins2017beta}
Irina Higgins, Loic Matthey, Arka Pal, Christopher Burgess, Xavier Glorot,
  Matthew Botvinick, Shakir Mohamed, and Alexander Lerchner.
\newblock {$\beta$-VAE}: Learning basic visual concepts with a constrained
  variational framework.
\newblock In {\em International Conference on Learning Representations}, 2017.

\bibitem{hu2018disentangling}
Qiyang Hu, Attila Szab{\'o}, Tiziano Portenier, Paolo Favaro, and Matthias
  Zwicker.
\newblock Disentangling factors of variation by mixing them.
\newblock In {\em IEEE Conference on Computer Vision and Pattern Recognition},
  pages 3399--3407, 2018.

\bibitem{huang2008labeled}
Gary~B Huang, Marwan Mattar, Tamara Berg, and Eric Learned-Miller.
\newblock Labeled faces in the wild: A database forstudying face recognition in
  unconstrained environments.
\newblock 2008.

\bibitem{huang2017beyond}
Rui Huang, Shu Zhang, Tianyu Li, and Ran He.
\newblock Beyond face rotation: Global and local perception gan for
  photorealistic and identity preserving frontal view synthesis.
\newblock In {\em IEEE International Conference on Computer Vision}, pages
  2439--2448, 2017.

\bibitem{huang2017arbitrary}
Xun Huang and Serge Belongie.
\newblock Arbitrary style transfer in real-time with adaptive instance
  normalization.
\newblock In {\em IEEE International Conference on Computer Vision}, pages
  1501--1510, 2017.

\bibitem{karras2019style}
Tero Karras, Samuli Laine, and Timo Aila.
\newblock A style-based generator architecture for generative adversarial
  networks.
\newblock In {\em IEEE Conference on Computer Vision and Pattern Recognition},
  pages 4401--4410, 2019.

\bibitem{kim2018deep}
Hyeongwoo Kim, Pablo Carrido, Ayush Tewari, Weipeng Xu, Justus Thies, Matthias
  Niessner, Patrick P{\'e}rez, Christian Richardt, Michael Zollh{\"o}fer, and
  Christian Theobalt.
\newblock Deep video portraits.
\newblock {\em ACM Transactions on Graphics}, 37(4):163, 2018.

\bibitem{kim2018disentangling}
Hyunjik Kim and Andriy Mnih.
\newblock Disentangling by factorising.
\newblock In {\em International Conference on Machine Learning}, 2018.

\bibitem{kulkarni2015deep}
Tejas~D Kulkarni, William~F Whitney, Pushmeet Kohli, and Josh Tenenbaum.
\newblock Deep convolutional inverse graphics network.
\newblock In {\em Advances in Neural Information Processing Systems}, pages
  2539--2547, 2015.

\bibitem{li2018closed}
Yijun Li, Ming-Yu Liu, Xueting Li, Ming-Hsuan Yang, and Jan Kautz.
\newblock A closed-form solution to photorealistic image stylization.
\newblock In {\em European Conference on Computer Vision (ECCV)}, pages
  453--468, 2018.

\bibitem{lin2019face}
Jinpeng Lin, Hao Yang, Dong Chen, Ming Zeng, Fang Wen, and Lu Yuan.
\newblock Face parsing with roi tanh-warping.
\newblock In {\em IEEE Conference on Computer Vision and Pattern Recognition},
  pages 5654--5663, 2019.

\bibitem{lin2019infogan}
Zinan Lin, Kiran~Koshy Thekumparampil, Giulia Fanti, and Sewoong Oh.
\newblock Infogan-cr: Disentangling generative adversarial networks with
  contrastive regularizers.
\newblock {\em arXiv preprint arXiv:1906.06034}, 2019.

\bibitem{pmlr-v97-locatello19a}
Francesco Locatello, Stefan Bauer, Mario Lucic, Gunnar Raetsch, Sylvain Gelly,
  Bernhard Sch{\"o}lkopf, and Olivier Bachem.
\newblock Challenging common assumptions in the unsupervised learning of
  disentangled representations.
\newblock In {\em International Conference on Machine Learning}, volume~97,
  pages 4114--4124, 2019.

\bibitem{mirza2014conditional}
Mehdi Mirza and Simon Osindero.
\newblock Conditional generative adversarial nets.
\newblock {\em arXiv preprint arXiv:1411.1784}, 2014.

\bibitem{nguyen2019hologan}
Thu Nguyen-Phuoc, Chuan Li, Lucas Theis, Christian Richardt, and Yong-Liang
  Yang.
\newblock {HoloGAN}: Unsupervised learning of 3d representations from natural
  images.
\newblock In {\em IEEE Conference on Computer Vision and Pattern Recognition},
  2019.

\bibitem{paysan20093d}
Pascal Paysan, Reinhard Knothe, Brian Amberg, Sami Romdhani, and Thomas Vetter.
\newblock A 3d face model for pose and illumination invariant face recognition.
\newblock In {\em IEEE International Conference on Advanced Video and Signal
  Based Surveillance}, pages 296--301, 2009.

\bibitem{perarnau2016icgan}
Guim Perarnau, Joost Van~de Weijer, Bogdan Raducanu, and Jose~M. Alvarez.
\newblock Invertible conditional gans for image editing.
\newblock In {\em Advances in Neural Information Processing Systems Workshop on
  Adversarial Training}, 2016.

\bibitem{piao2019semi}
Jingtan Piao, Chen Qian, and Hongsheng Li.
\newblock Semi-supervised monocular 3d face reconstruction with end-to-end
  shape-preserved domain transfer.
\newblock In {\em IEEE International Conference on Computer Vision}, pages
  9398--9407, 2019.

\bibitem{pumarola2018ganimation}
Albert Pumarola, Antonio Agudo, Aleix~M Martinez, Alberto Sanfeliu, and
  Francesc Moreno-Noguer.
\newblock Ganimation: Anatomically-aware facial animation from a single image.
\newblock In {\em European Conference on Computer Vision}, pages 818--833,
  2018.

\bibitem{qian2019unsupervised}
Yichen Qian, Weihong Deng, and Jiani Hu.
\newblock Unsupervised face normalization with extreme pose and expression in
  the wild.
\newblock In {\em IEEE Conference on Computer Vision and Pattern Recognition},
  pages 9851--9858, 2019.

\bibitem{ramamoorthi2001efficient}
Ravi Ramamoorthi and Pat Hanrahan.
\newblock An efficient representation for irradiance environment maps.
\newblock In {\em SIGGRAPH}, pages 497--500, 2001.

\bibitem{reed2014learning}
Scott Reed, Kihyuk Sohn, Yuting Zhang, and Honglak Lee.
\newblock Learning to disentangle factors of variation with manifold
  interaction.
\newblock In {\em International Conference on Machine Learning}, pages
  1431--1439, 2014.

\bibitem{sengupta2018sfsnet}
Soumyadip Sengupta, Angjoo Kanazawa, Carlos~D Castillo, and David~W Jacobs.
\newblock Sfsnet: Learning shape, reflectance and illuminance of facesin the
  wild'.
\newblock In {\em IEEE Conference on Computer Vision and Pattern Recognition},
  pages 6296--6305, 2018.

\bibitem{shen2019interpreting}
Yujun Shen, Jinjin Gu, Xiaoou Tang, and Bolei Zhou.
\newblock Interpreting the latent space of gans for semantic face editing.
\newblock {\em arXiv preprint arXiv:1907.10786}, 2019.

\bibitem{shen2018faceid}
Yujun Shen, Ping Luo, Junjie Yan, Xiaogang Wang, and Xiaoou Tang.
\newblock {FaceID-GAN}: Learning a symmetry three-player gan for
  identity-preserving face synthesis.
\newblock In {\em IEEE Conference on Computer Vision and Pattern Recognition},
  pages 821--830, 2018.

\bibitem{shen2018facefeat}
Yujun Shen, Bolei Zhou, Ping Luo, and Xiaoou Tang.
\newblock Facefeat-gan: a two-stage approach for identity-preserving face
  synthesis.
\newblock {\em arXiv preprint arXiv:1812.01288}, 2018.

\bibitem{song2018geometry}
Lingxiao Song, Zhihe Lu, Ran He, Zhenan Sun, and Tieniu Tan.
\newblock Geometry guided adversarial facial expression synthesis.
\newblock In {\em ACM Multimedia Conference on Multimedia Conference}, pages
  627--635, 2018.

\bibitem{sun2019single}
Tiancheng Sun, Jonathan~T Barron, Yun-Ta Tsai, Zexiang Xu, Xueming Yu, Graham
  Fyffe, Christoph Rhemann, Jay Busch, Paul Debevec, and Ravi Ramamoorthi.
\newblock Single image portrait relighting.
\newblock {\em ACM Transactions on Graphics}, 38(4):79, 2019.

\bibitem{tenenbaum2000separating}
Joshua~B Tenenbaum and William~T Freeman.
\newblock Separating style and content with bilinear models.
\newblock {\em Neural computation}, 12(6):1247--1283, 2000.

\bibitem{tran2017disentangled}
Luan Tran, Xi Yin, and Xiaoming Liu.
\newblock Disentangled representation learning gan for pose-invariant face
  recognition.
\newblock In {\em IEEE Conference on Computer Vision and Pattern Recognition},
  pages 1415--1424, 2017.

\bibitem{whitney2016understanding}
William~F Whitney, Michael Chang, Tejas Kulkarni, and Joshua~B Tenenbaum.
\newblock Understanding visual concepts with continuation learning.
\newblock {\em arXiv preprint arXiv:1602.06822}, 2016.

\bibitem{xiao2018elegant}
Taihong Xiao, Jiapeng Hong, and Jinwen Ma.
\newblock Elegant: Exchanging latent encodings with gan for transferring
  multiple face attributes.
\newblock In {\em European Conference on Computer Vision}, pages 168--184,
  2018.

\bibitem{xu2020portrait}
Sicheng Xu, Jiaolong Yang, Dong Chen, Fang Wen, Yu Deng, Yunde Jia, and Xin
  Tong.
\newblock Deep 3d portrait from a single image.
\newblock In {\em IEEE Conference on Computer Vision and Pattern Recognition},
  2020.

\bibitem{yang2017neural}
Jiaolong Yang, Peiran Ren, Dongqing Zhang, Dong Chen, Fang Wen, Hongdong Li,
  and Gang Hua.
\newblock Neural aggregation network for video face recognition.
\newblock In {\em IEEE Conference on Computer Vision and Pattern Recognition},
  pages 4362--4371, 2017.

\bibitem{yin2017towards}
Xi Yin, Xiang Yu, Kihyuk Sohn, Xiaoming Liu, and Manmohan Chandraker.
\newblock Towards large-pose face frontalization in the wild.
\newblock In {\em IEEE International Conference on Computer Vision}, pages
  3990--3999, 2017.

\bibitem{zhao2018towards}
Jian Zhao, Yu Cheng, Yan Xu, Lin Xiong, Jianshu Li, Fang Zhao, Karlekar
  Jayashree, Sugiri Pranata, Shengmei Shen, Junliang Xing, et~al.
\newblock Towards pose invariant face recognition in the wild.
\newblock In {\em IEEE conference on computer vision and pattern recognition},
  pages 2207--2216, 2018.

\bibitem{zhou2019deep}
Hao Zhou, Sunil Hadap, Kalyan Sunkavalli, and David~W Jacobs.
\newblock Deep single-image portrait relighting.
\newblock In {\em IEEE International Conference on Computer Vision}, pages
  7194--7202, 2019.

\end{thebibliography}
}
\includepdf[pages=1]{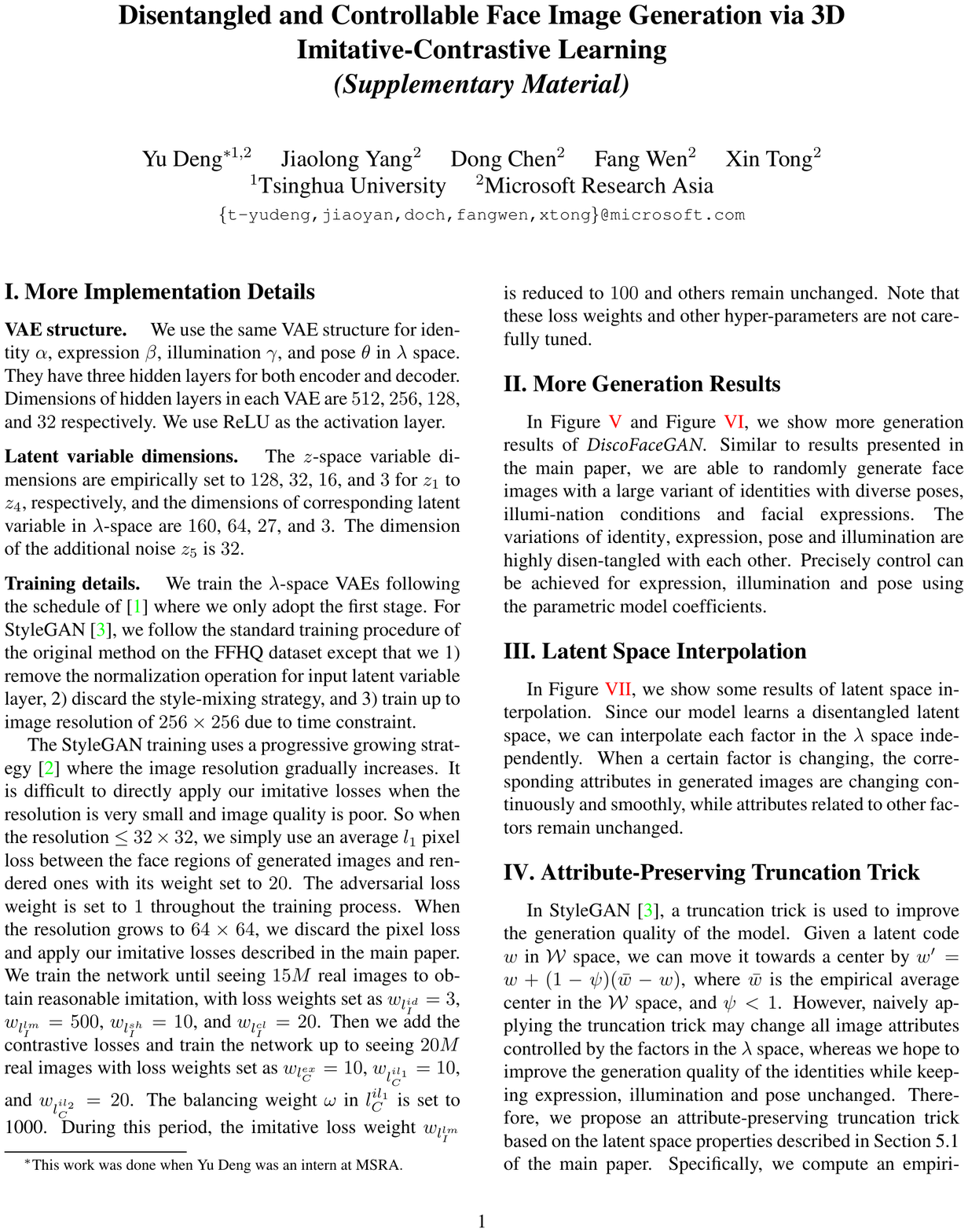}
\includepdf[pages=2]{supplementary.pdf}
\includepdf[pages=3]{supplementary.pdf}
\includepdf[pages=4]{supplementary.pdf}
\includepdf[pages=5]{supplementary.pdf}
\includepdf[pages=6]{supplementary.pdf}
\includepdf[pages=7]{supplementary.pdf}
\end{document}